\pdfoutput=1

\documentclass[11pt]{article}
\usepackage{setspace}
\usepackage[dvipsnames]{xcolor}
\usepackage{graphicx}
\usepackage{tabularx}
\usepackage{afterpage}
\usepackage{booktabs}
\usepackage{amssymb}
\usepackage{amsmath}
\usepackage{enumitem}
\usepackage{hyperref}

\usepackage{acl}

\usepackage{times}
\usepackage{latexsym}
\usepackage{xspace}
\usepackage{multirow}
\usepackage{makecell}
\usepackage{subcaption}
\usepackage[T1]{fontenc}

\usepackage[table]{xcolor}
\usepackage{colortbl} 

\newif\ifshowcomments

\newcommand{\cut}[1]{}
\newcommand{\xhdr}[1]{{\noindent\bfseries #1.}}

\newcommand{\isalert}{g\xspace}

\newcommand{\pderail}{\mathcal{P}(\mathrm{derailment}\,\big|\, u_1,u_2,\ldots,u_k)\xspace}
\newcommand{\pderailsim}{\mathcal{P}(\mathrm{derailment}\,\big|\, u_1,...,u_k, \mathbf{u_{k+1}^\mathbf{\mathrm{sim}_i}})\xspace}

\usepackage[utf8]{inputenc}

\usepackage{microtype}

\usepackage{inconsolata}
\usepackage{float}

\title{Wait! There’s a Way Out: \\ A Decision Mechanism for Forecasting Conversational Derailment}

\author{Laerdon Kim, Vivian Nguyen, Cristian Danescu-Niculescu-Mizil \\
Cornell University \\
\texttt{\{lyk25, vn72\}@cornell.edu} \hspace{8pt} \texttt{cristian@cs.cornell.edu
}
}

\definecolor{calm_green}{RGB}{125, 181, 110}
\definecolor{awry_red}{RGB}{201, 81, 81}
\begin{document}

\maketitle

\begin{abstract}

Forecasting conversational derailment is the task of predicting, as the conversation unfolds,  whether it will eventually derail into personal attacks.
Since forecasting models operate in an online fashion, they must decide whether to "trigger" an alert after each utterance---for example, to notify participants or a moderator that the conversation is at risk of derailing. 
Existing approaches make this decision solely based on the estimated likelihood of derailment given the preceding utterances, implicitly assuming that the conversation’s 
future trajectory is fixed.
As a result, they ignore the possibility of future recovery and incur an unnecessarily high rate of false positives.

In this work we propose a method for decoupling the decision to trigger from the derailment likelihood estimation.
Our approach is inspired by the first human baseline on this task, which shows that humans achieve dramatically lower false positive rates by selectively deferring their decision to trigger when they anticipate that tension is likely to subside.
We operationalize this insight with a deferral mechanism that uses forward-looking simulations to assess whether a tense moment admits plausible paths to recovery.
Incorporating this mechanism into a state-of-the-art forecasting model substantially reduces false positives 
without sacrificing forecasting accuracy.
More broadly, this work highlights the value of treating decision making as a first-class component of forecasting systems.
\end{abstract}

\section{Introduction}
\label{sec:intro}
\begin{quote}
    "The best way to predict the future is to invent it." -- Alan Kay
\end{quote}

\begin{figure*}[t]
    \centering
    \includegraphics[width=0.9\linewidth]{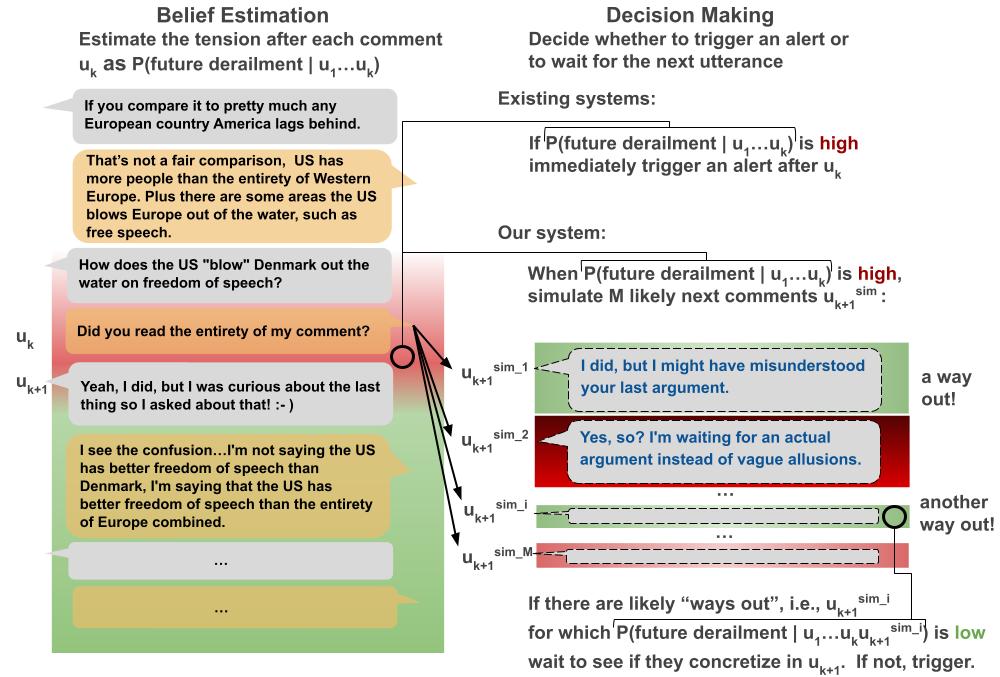}
    \caption{Current systems for forecasting conversational derailment conflate tension estimation with the decision to trigger an alert. 
    As illustrated in this example, this can lead to false positives.
    We devise a mechanism that defers triggering when it anticipates a possible de-escalation (via forward-looking simulations).
    Background colors indicate the probability of derailment calculated based on the context up to that utterance (from \textcolor{calm_green}{low} to \textcolor{awry_red}{high}).}%
    \label{fig:delay-intro-figure}
\end{figure*}

Conversational forecasting is the task of predicting whether a conversational event---such as a personal attack \cite{chang_trouble_2019},  disengagement \cite{nguyen_hanging_2025}, or prosocial behavior \cite{bao_conversations_2021}---will \textit{eventually} occur in the conversation.
The ability to make such predictions was recognized as a key component of proactive conversational support systems \cite{jurgens_just_2019,korre_evaluation_2025},
and its potential was demonstrated via academic user studies \cite{chang_thread_2022,schluger_proactive_2022}, and real-world  product deployments \cite{li_using_2022}.

One of the central challenges in forecasting---distinguishing it from traditional static classification tasks---is the ``unknown horizon'' of the target event \cite{chang_thread_2022}.
Take the example of forecasting conversational derailment, which is the focus of this work: a personal attack can occur at any time, requiring a forecasting system to decide after every comment whether to trigger an alert or to wait for the conversation to further develop.
Early triggering can lead to spurious alarms driven by limited information, whereas delayed triggering risks missing the opportunity for timely intervention.
Consequently, effective forecasting models must separate \textit{belief estimation} from \textit{decision-making}: beyond estimating the current risk of derailment  (\autoref{fig:delay-intro-figure} left), they must decide whether the evidence observed so far warrants triggering an alarm or whether it is preferable to wait for more information (\autoref{fig:delay-intro-figure} right).

However, all existing models conflate belief estimation with decision-making by triggering alerts via a fixed threshold on the estimated probability of derailment, regardless of whether this probability is calculated via hierarchical recurrent neural networks \cite{chang_trouble_2019}, graph convolutional networks \cite{altarawneh_conversation_2023}, transformer architectures \cite{kementchedjhieva_dynamic_2021}, hierarchical transformer architectures
\cite{yuan_conversation_2023}, or LLM prompting \cite{olpadkar_can_2025}.
In effect, these models function as classifiers that estimate the level of conversational tension up to the current moment and trigger an alert once a predetermined threshold is crossed (in \autoref{fig:delay-intro-figure} an existing model would trigger right after $u_k$).
As a result, such models are blind to the possibility that conversational tension may later subside (a possibility exemplified in \autoref{fig:delay-intro-figure} starting with $u_{k+1}$).
Consequently, current forecasting systems cannot reason about the value of waiting for additional evidence before triggering.

Beyond being a theoretical limitation, the absence of a separate decision-making component has important practical consequences for forecasting systems and their real-world deployment.
In particular, the inability to account for the possibility that conversational tension may later resolve leads directly to spurious alerts (i.e., false positives),
which end users identify as the most significant shortcoming of such systems, with 62\% reporting false positives as a common issue \citep{chang_thread_2022}.

In this work, we propose the first approach that decouples the triggering decision from the derailment likelihood estimation.
We begin by examining how humans solve the forecasting task and, in the process, establish the first human baseline.
We find that humans achieve a false positive rate that is less than half of the state-of-the-art forecasting model by waiting longer before triggering an alert.
Crucially, this deferral of the decision to trigger is selective: humans appear to anticipate when conversational tension is likely to subside.

To operationalize this insight, we devise a simulation-based method for distinguishing tense moments after which a recovery is plausible---that is, there is a ``way out'' (\autoref{fig:delay-intro-figure}, "Our system").
We implement a deferral mechanism which acts cautiously (only) in these tense moments, waiting for the next utterance rather than immediately triggering an alert. 
Integrating this mechanism on top of a state-of-the-art forecasting model yields a dramatic reduction in false positives without sacrificing overall accuracy.

In summary, in this work we:
\begin{itemize}
    \item conceptualize and disentangle two components of conversational forecasting that have previously been conflated: belief estimation and decision-making;
    \item introduce the first human baseline for forecasting conversational derailment, showing that humans make more effective triggering decisions to achieve substantially lower false positive rates than state-of-the-art models;
    \item propose a decision mechanism inspired by this insight that narrows this false positive gap without sacrificing overall accuracy.
\end{itemize}

More broadly, this work illustrates the value of disentangling the decision to trigger from belief estimation in forecasting systems, opening the door to more sophisticated and effective trigger policies.
To encourage further progress in this direction, we publicly release our code in a modular form that explicitly separates the two components, rendering it forward-compatible with alternative tension estimation models and decision mechanisms.\footnote{Code and data released as part of \href{https://convokit.cornell.edu/}{ConvoKit}.}

\section{Background and Related Work}
\label{sec:bgrelated}

\xhdr{Conversations Gone Awry} The development of conversational forecasting models has been catalyzed by the introduction of the Conversations Gone Awry (CGA) task \cite{zhang_conversations_2018,chang_thread_2022,tran_conversations_2025}.
In this task, the event to be forecast is the occurrence of a personal attack.
Accordingly, a model processes the conversation one utterance at a time, with the goal of \textit{triggering} an alert only if it determines that the conversation is likely to culminate in a personal attack.
For a model to be successful, it must either trigger an alert \textit{before} a personal attack occurs (when one exists), or correctly refrain from triggering and allow the conversation to reach its end when no personal attack occurs.

The task includes two datasets.  
CGA-CMV was introduced by \citet{chang_trouble_2019} and later expanded by \citet{tran_conversations_2025} to include $20{,}576$ conversations collected from the Change My View subreddit.
Labels are derived from moderator-marked violations of Rule~2 of Change My View: ``Don’t be rude or hostile to other users.''  
CGA-WIKI \cite{zhang_conversations_2018} is a  much smaller set of $4{,}188$ conversations between Wikipedia editors, human-labeled by crowdworkers as whether they end in personal attacks, and filtered to exclude any ``rude, insulting, or disrespectful'' comments prior to the final turn.
As a result of this additional filtering, CGA-WIKI is less naturalistic: it is unlikely to contain conversations exhibiting de-escalating or recovery trajectories.
Therefore in this work we focus on CGA-CMV, but report the results on CGA-WIKI for completeness in \autoref{appendix:wiki}.

\xhdr{Unknown horizon}
The online formulation of the task gives rise to the aforementioned ``unknown horizon'' challenge inherent in conversational forecasting \cite{chang_trouble_2019}.  
Because a personal attack---if it occurs---may arise at any point in the interaction, the model does not know in advance when it must act.
Instead, it must repeatedly decide, after each utterance, whether to act or to wait, without knowing whether additional evidence will arrive or whether the conversation is about to conclude.  
As a result, the model must balance the risk of acting too early, based on insufficient information, against the risk of acting too late, when intervention may no longer be effective.

Although this challenge and inherent exploration-exploitation tradeoff were recognized when the task was first introduced and is reflected in the evaluation protocol---where forecasters which fail to intervene before the conversation derails are penalized---current forecasting systems lack explicit mechanisms for deciding when to act.
Instead, they effectively operate as traditional threshold-based classifiers, triggering alerts whenever the estimated probability of derailment crosses a fixed cutoff.

\xhdr{Forecasting formalism: belief estimation}
Formally, the belief estimation component of a forecaster continuously assesses the probability of future derailment after the $k$-th utterance $u_k$, conditioned on the preceding context: $\pderail$.
A variety of models have been proposed to estimate this probability, drawing on architectures such as hierarchical recurrent neural networks \cite{chang_trouble_2019}, graph convolutional networks \cite{altarawneh_conversation_2023}, transformers \cite{kementchedjhieva_dynamic_2021}, hierarchical transformers \cite{yuan_conversation_2023}, and decoder-based generative large language models \cite{olpadkar_can_2025,tran_conversations_2025}.
These architectures were recently compared using the official public benchmark for the CGA task \cite{tran_conversations_2025}.

Across both the CGA-CMV and CGA-WIKI sections of the benchmark, the best-performing model is Gemma2 9B \cite{teamGemma2Improving2024},
which we therefore treat as the state-of-the-art (SOTA) for this task.
Such decoder-only models are fine-tuned on the training data;
 the probability of derailment $\pderail$ is retrieved by prompting the model to answer whether the conversation will derail after $u_k$,  sampling the ``Yes'' and ``No'' logits, and softmaxing the ``Yes'' probability \cite{tran_conversations_2025}.

A key challenge in training 
forecaster models is that supervision is available only at the conversation level: each conversation is labeled solely by whether it ultimately ends in a personal attack, with no utterance-level supervision \cite{chang2024thesis}.
Accordingly, prior work trains models to predict the label of the final utterance $u_n$ using the preceding context $u_1,\ldots,u_{n-1}$---even though at test time models must produce a derailment probability after every utterance, not just the last.
While this strategy has been shown to outperform training on partial conversations 
\citep{altarawneh_conversation_2023}, 
it structurally biases models toward detecting immediate tension, rather than learning about longer-term conversational dynamics.
As a result, these models cannot account for more complex conversational trajectories, such as moments that are (or appear to be) tense but are followed by a reduction in perceived tension.
Such recoveries can naturally arise through de-escalation (as illustrated in Figure~\ref{fig:delay-intro-figure}) or through the repair of a misunderstanding or misperception (e.g., a comment that initially appears sarcastic but is later clarified as earnest; \citet{tsai_leveraging_2024}, \citet{drew_open_1997}).

\xhdr{Forecasting formalism: decision-making}
After every utterance $u_k$, the {decision component} of a forecasting system must choose whether to trigger an alert ($\isalert_k = 1$) or to wait for the next utterance ($\isalert_k = 0$).  
The responsibility of this component ends either when an alert is triggered---at which point the conversation is deemed to be ``going awry'' 
---or when the conversation reaches its end without an alert, in which case it is deemed to ``remain calm''. %
Current systems, including the SOTA, use a simple threshold based decision, triggering when the estimated tension surpasses a threshold $T$ fixed on the validation data:

\begin{equation}
 g_k := \mathbb{I}\{ \; \pderail > T \; \} 
\end{equation}

In this work, we treat decision-making as a first-class component of forecasting models and propose the first framework in which triggering decisions are informed not only by the estimated derailment probability, but also by forward-looking signals about how the conversation may evolve, including the plausibility of a future recovery.

\xhdr{Evaluation metrics}
In forecasting tasks, standard error types link utterance-level predictions to conversation-level labels \cite{chang_trouble_2019}.
A conversation that ends with a personal attack is a true positive if the model triggers before the conversation ends ($\isalert_k = 1$ for some $k<n$), and a false negative otherwise ($\isalert_k = 0$ for all $k<n$).
Conversely, a conversation that does not end with a personal attack is a false positive if the model triggers at any utterance, and a true negative otherwise.
Prior work uses these error types to define standard evaluation metrics for comparing forecasting models. 
Accuracy is often preferred (e.g., for selecting the triggering threshold) over F1 and false-positive rate (FPR), as it accounts for all error types.
In addition, the ``horizon'' ($H$) is a forecasting-specific metric that measures how many utterances before an actual personal attack the model triggers an alert (averaging across all true positives).

\xhdr{Forecasting and simulation}
Since conversational forecasting concerns predicting future conversational events, it is natural to consider whether the generative capabilities of large language models can be used to “peek” into possible future trajectories.
In fact, the first conversational forecasting systems were built on top of a generative framework (Hierarchical Recurrent Encoder–Decoder), replacing the generation layer with a prediction head \cite{chang_trouble_2019}.
More recently, \citet{zhang_forecasting_2025} evaluated the utility of modern large language model simulations in a \textit{static} variant of the forecasting task.
In this static variant, however, the forecaster always makes its prediction at a predetermined point in the conversation---namely, immediately before the final comment, assuming (an unrealistic) prior knowledge of when the conversation is about to end---thereby removing the need for an explicit decision-making component.
This work inspires one of our baselines in the full version of the task.

In the context of mental health crisis counseling conversations, \citet{nguyen_hanging_2025} shows the value of combining next-utterance simulation with forecasting in determining when a moment is ``pivotal''. 
Our proposed method also uses next-utterance simulations, but with the goal of deciding whether there is informational value in deferring the decision to trigger an alarm.

\section{Human Experiment}
\label{sec:human}

To inform the design of the decision component in conversational forecasting systems, we turn to observing how humans solve the task.
We analyze the human triggering decisions and directly compare them to those of the SOTA model, with the goal of identifying systematic differences that can inspire more effective mechanisms for deciding when (not) to trigger an alert.

Despite its prominence, the CGA task has so far lacked a human baseline.
This gap stems in large part from the online nature of the task: unlike standard classification settings, forecasting conversational derailment requires making sequential decisions, where annotators must decide when to act rather than simply what label to assign.
Moreover, because the task involves predicting future events, annotation quality cannot be directly verified at each decision point, making it difficult to rely on standard crowdsourcing pipelines.
To address this gap, we devise a human experiment that places participants in this online forecasting setting, enabling a direct comparison between human and model decision-making.

In order to match the forecasting environment while incentivizing participants, we follow previous work in ``gamifying'' the task \cite{vannella_validating_2014}. 
Under this formulation, participants are presented with conversations one comment at a time and must choose either to trigger an alert---if they believe the conversation is getting out of hand---or to reveal the next comment. 
Their objective is to reveal all utterances in non-derailing conversations, while triggering an alert in derailing conversations \textit{before} revealing the personal attack itself. 
A snapshot of the experiment's interface is included in \autoref{appendix:humanexpscreencap}.
To incentivize participants, we allow them to see whether they answered correctly or not for each conversation, and to compare total scores among themselves after each round.

The experiment involves three rounds.
The first one is a warmup round consisting of four hand-picked conversations, to allow the participants to become familiar with the task and ask questions. 
The results of this round are discarded from the analysis.
Then we run two main rounds: in each round, each participant goes through 10 conversations.
These conversations are randomly selected from CGA-CMV-Large such that half conclude with a personal attack and the other half remain civil.
Given this setup, a random baseline achieves an accuracy of $50\%$ like in the original 
CGA task.

To ensure comparability between the two experimental rounds, we use the same
set of conversations for both rounds, only changing assignment of conversations to participants between the rounds.
In this way, we ensure that no participant encounters the same conversation twice.
This results in all conversations receiving annotation from two unique participants: one in the first main round and one in the second.
We recruited nine volunteers, each of whom completed all rounds of the game, spending an average of $13$ minutes per main round. 
Due to the random assignment of conversations, some were repeated across participants, resulting in a total of $84$ unique conversations.

\begin{table}[!t]
    \centering
   \resizebox{\linewidth}{!}
    {
    \begin{tabular}{l|cccccc}
    \toprule
         Method & Acc $\uparrow$ & FPR $\downarrow$ & P $\uparrow$ & R $\uparrow$ & F1 $\uparrow$ &  H  \tabularnewline
        \midrule
        Human rnd-1 & 62.2 & 24.4 & 67.8 & 48.9 & 54.6 & 2.6 \tabularnewline
         \phantom{Human} rnd-2 & \textbf{70.0} & \textbf{15.6}  & 75.9 & 55.6 & 63.9 & 2.1 \tabularnewline
        \midrule
        SOTA & \textbf{70.0} & 36.2 & \textbf{67.9} & \textbf{76.2} & \textbf{71.8} & 3.3 \tabularnewline
    \bottomrule
    \end{tabular}}
    \caption{
        Human performance micro-averaged across 84 randomly selected conversations from CGA-CMV-Large,  compared with the SOTA performance on the same sample (averaged across 5 random seeds).
        We report accuracy (Acc), precision (P), recall (R), F1, false positive rate (FPR), and horizon (H).
    }
    \label{tab:survey-cmv_large}
\end{table}

\xhdr{Human performance} Table~\ref{tab:survey-cmv_large} shows that after the second round, human participants achieve accuracy comparable to that of the state-of-the-art (SOTA) model.
Importantly, humans exhibit a substantially lower false positive rate---less than half that of the SOTA model---albeit at the cost of recall.
This more cautious strategy is further reflected in a shorter mean horizon,\footnote{As discussed in \autoref{sec:bgrelated}, the horizon is a forecasting-specific metric introduced by \citet{chang_thread_2022} and adopted by subsequent work; it measures how many comments before the personal attack an alert is triggered.} indicating that participants tend to wait longer before guessing that a conversation will derail.

Notably, participants improve from the first round to the second, achieving higher accuracy while simultaneously becoming more cautious, as evidenced by decreases in both false positive rate and horizon.
This pattern suggests that with experience, humans can calibrate their decision-making, refining their intuitions about conversational trajectories and learning to avoid false positives.

\xhdr{Comparing human and SOTA decisions}
To better understand the human decision process---and in particular how humans achieve such a substantial reduction in false positive rate---we directly compare human triggering behavior with that of the SOTA model.
A natural hypothesis is that humans are simply more cautious than the model, requiring higher levels of conversational tension before triggering an alert.
To test this hypothesis, we compare the model-inferred tension (i.e., $\pderail$)\footnote{For ease of exposition, in what follows we use ``tension'' to refer to the probability of derailment estimated by the SOTA model, following  the discussion about the training procedure of these models in \autoref{sec:bgrelated}. We acknowledge this as a necessary approximation used for exploratory purposes, since there are no labels of human-perceived tension.} at the moment humans trigger 
an alert with the level of tension at which the SOTA model triggers.
We find evidence against this hypothesis: humans trigger, on average, at \textit{lower} levels of estimated tension ($\mathcal{P}=0.61$) than the model ($\mathcal{P}=0.72$).

Since humans do not appear to be more cautious in general, their lower false positive rate may instead stem from \textit{selectively} exercising caution in situations where they anticipate that tension will subside.
To explore this possibility, we define the immediate decrease in tension following utterance $k$ as:
\begin{equation} \begin{aligned} \mathcal{D}_k :=\;& \mathcal{P}(\mathrm{derailment}\,\big|\, u_1,u_2,\ldots,u_{k}) \\ &- \mathcal{P}(\mathrm{derailment}\,\big|\, u_1,u_2,\ldots,u_{k+1}). \end{aligned} \end{equation}
\noindent with a strictly positive value indicating a decrease in tension.\footnote{We discard cases in which $u_{k+1}$ is the to-be-forecasted personal attack or the last comment of the conversation.} 

We find that a large fraction (61\%) of moments in which SOTA triggers are actually followed by a decrease in tension ($\mathcal{D}_k>0$).
In comparison, humans are much less likely to trigger in moments that are followed by an immediate decrease in tension (24\% of moments in which humans trigger have $\mathcal{D}_k > 0$).

If we focus on moments in which humans successfully avert a false positive---specifically, utterances where the SOTA triggers a false alert but humans choose to wait and allow the conversation to continue---we find in 78.6\% of these cases, tension subsequently decreases, 
a  higher proportion than observed for SOTA false positive triggers in general (71.6\%).
Therefore, even in moments where derailment appears plausible to the SOTA model, humans selectively refrain from triggering when they sense that the interaction is likely to recover.

\section{Method}
\label{sec:method}

Humans appear to defer the decision to trigger an alert in moments of high but transient tension.
Motivated by this observation, our goal is to design a decision mechanism that explicitly incorporates the ability to anticipate possible recoveries into the forecaster's decision-making process.
We introduce a decision component that directly accounts for possible future conversational dynamics.
This enables the forecaster to distinguish between tense moments that are likely to persist and those that appear recoverable.

We take a simulation-based approach to operationalize this intuition.
At each tense moment in a conversation, we simulate plausible next replies in order to estimate whether an observed spike in tension is likely to persist or to subside (as illustrated in Figure~\ref{fig:delay-intro-figure}).
Formally, let $k$ denote a point in the conversation that the belief estimation component deems tense, i.e., $\pderail > T$.
These are moments in which a traditional forecasting model without a distinct decision-making component would trigger an alert.
We simulate $M$ different possibilities for the next utterance, $u_{k+1}^{\mathrm{sim}_i}$, for $i = 1,\ldots,M$.
For each such simulated continuation, we again apply the belief estimation component to update the estimated derailment probability after the simulated utterance, and record the corresponding threshold-based triggering decision:
\begin{equation}
g_{k+1}^{\mathrm{sim}_i} = \mathbb{I}\{\pderailsim > T \} 
\end{equation}

We use this information to devise a decision-making mechanism that defers the trigger decision in tense moments when simulated continuations indicate plausible paths to conversational recovery:

\begin{equation}
    \isalert_k^{\textrm{+}} = 
    \begin{cases}
        1 & \text{if } {\mathcal{P}(\mathrm{derailment}\,\big|\, u_1,\ldots,u_k)\xspace} > T  \\ & \text{ and }
        \{ M - \sum_i g_{k+1}^{\mathrm{sim}_i} \} \le \tau 
        \\ 0 & \text{otherwise}
        
    \end{cases}
\label{eq:trigger-delay}
\end{equation}

\noindent where $\tau$ is a parameter that controls how much simulated evidence of recovery---i.e., how many $i$ satisfy $g_{k+1}^{\mathrm{sim}_i}=0$---is required to defer the decision to trigger, rather than trigger immediately as the SOTA forecaster would do.
In order for an awry prediction to persist and become a trigger, under this mechanism, the simulated replies which are calm must not exceed $\tau$.
We simulate $M=10$ next utterances for each tense moment, and require that more than two thirds of the simulations ($\tau=7$)
 indicate recovery in order to defer a threshold-based triggering decision.
 We list further implementation details and model parameters in \autoref{appendix:implementation}.

\section{Results}
\label{sec:results}

\begin{table*}[t]
\centering
{
\begin{tabular}{lcccccc}
\toprule
Method & Acc $\uparrow$ & FPR $\downarrow$ & P $\uparrow$ & R $\uparrow$ & F1 $\uparrow$ & {H} \\
\midrule
SOTA         & \textbf{70.9} & 34.3 & 69.1 & 76.1 & \textbf{72.3} & 3.9 \\
\textbf{+ selective deferral} & \textbf{70.9} & \textbf{\textcolor{red}{26.7}} & \textbf{72.1} & 68.4 & 70.2 & 3.8 \\
 + random deferral & 69.4 & 30.2 & 70.0 & 69.0 & 69.2 & 3.8 \\
\midrule
+ simulation (average)    & 70.2 & 36.2 & 68.1 & 76.6 & 72.0 & 4.0 \\
+ simulation (majority) & 70.0 & 36.9 & 67.7 & \textbf{76.7} & 71.8 & 4.0 \\
\midrule
\midrule
oracle threshold  & 70.0 & 26.7 & 71.5 & 66.8 & 69.0 & 3.7 \\
\bottomrule
\end{tabular}
}
\caption{Performance on CGA-CMV-Large, averaged across 5 random seeds. Selective deferral drastically reduces the FPR while maintaining accuracy. Even when selecting an oracle threshold on the test set, SOTA cannot achieve as good of a precision-recall tradeoff without a deferral mechanism.}
\label{tab:cmv_avg_perf}
\end{table*}

Here, we focus on quantifying the effect of adding an explicit decision-making component to the state-of-the-art model on the CGA-CMV-Large test set, deferring discussion of CGA-WIKI results to \autoref{appendix:wiki}.
We compare against several baseline systems to assess whether the observed gains can be attributed to a less informed deferral strategy, or merely to the use of next-utterance simulation itself.
We additionally construct an oracle system (with unrealistic access to the test set) to assess whether it is possible to achieve similar results by simply finding an ideal threshold, or whether a separate decision component is necessary.

\xhdr{Simulation baselines}
Inspired by \citet{zhang_forecasting_2025}, we directly use the simulated continuations $u_{k+1}^{\mathrm{sim}_i}$ to estimate the derailment probability $\pderailsim$.
We use two versions of this baseline: one that takes the average of the derailment probability across all simulated utterances, and one that takes the majority vote of $g_{k+1}^{\mathrm{sim}_i}$.
Both of these use a pre-determined trigger threshold, and allow us to verify whether the information in the simulated replies is sufficient in the absence of a separate decision mechanism.

\xhdr{Random deferral baseline}
We ablate the selective part of the deferral mechanism, which considers the number of calm simulations $(g^{sim_i}_{k+1} = 0)$ to decide whether to defer, by producing a baseline that defers a decision to trigger randomly
(maintaining the same likelihood of deferral as in the full system, as observed on the training data).\footnote{A similar result is obtained if  the likelihood of deferral is estimated on the test data. This result is included together with other baselines in \autoref{appendix:morebaselines}.}

\xhdr{Oracle threshold selection}
To obtain an upper bound on what the SOTA system can achieve without a separate decision mechanism, we consider an oracle version of the SOTA with the threshold $T$ tuned on the test data (and thus is unrealistic in practice).
To compare the precision-recall tradeoff, we pick a threshold that leads to a similar FPR to our full model.\footnote{Given the non-uniform distribution of derailment probabilities, it is not possible to perfectly match the FPR.}

The results in \autoref{tab:cmv_avg_perf} demonstrate that adding the deferral-based decision mechanism to the SOTA system yields a substantial reduction in false positive rate, effectively narrowing the gap between human performance without sacrificing overall accuracy.
None of the alternative baselines achieve comparably low false positive rates at similar accuracy levels.

\begin{figure*}[t]
\centering
\includegraphics[width=0.7\linewidth]{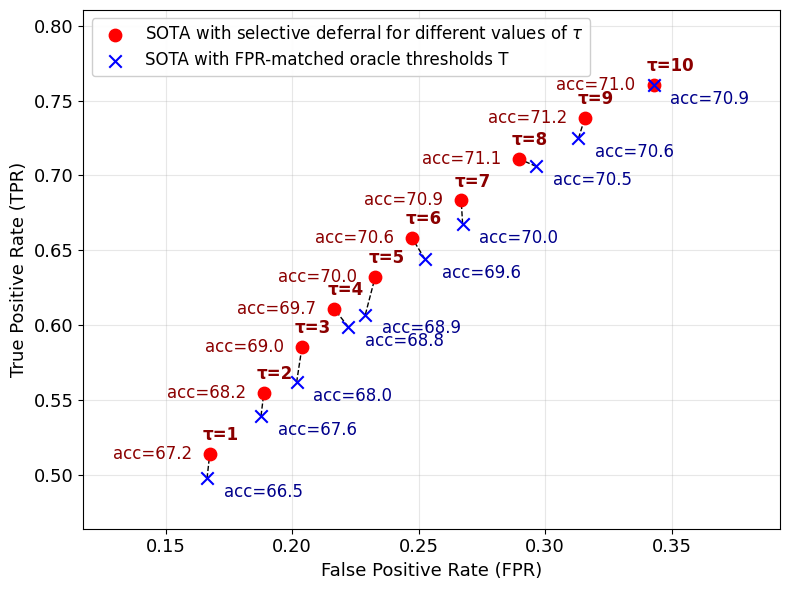}
\caption{Comparing precision-recall tradeoff on CGA-CMV-Large. For any choice of $\tau$, selective deferral yields a better recall (and better accuracy) than SOTA with the best possible threshold---as found on the test set---that matches that respective FPR. See \autoref{table:taumatching} in the Appendix for a comparison across all metrics.}
\label{fig:ROC}
\end{figure*}

Moreover, these improvements cannot be achieved by SOTA models without a separate decision mechanism. 
Even tuning the decision threshold on the test set yields a poorer precision–recall tradeoff (oracle threshold in \autoref{tab:cmv_avg_perf}). 
As shown in \autoref{fig:ROC}, this holds for all values of $\tau$: without the deferral mechanism, SOTA yields lower recall for each corresponding FPR.

Taken together, these findings suggest that the ability to selectively defer is key in achieving such reductions in false positive rates.
This result underscores the value of explicitly separating decision making from belief estimation in conversational forecasting, and motivates future work on richer and more flexible triggering mechanisms.

\section{Deferral Decisions}
\label{sec:qualitative}

\begin{table*}[t]
\centering
\begin{tabular}{p{3cm} p{12cm}}
\toprule
\textbf{Phenomenon} & \textbf{Example (distinguishing phrase in bold)} \\
\midrule

\multirow{1}{*}{\parbox{3cm}{Direct attribution of fault}}
& \textbf{The fact that} you don't want to talk [...] shows you can't defend [...] \\
& Why in the world would \textbf{you think that} beliefs aren't a choice? \\
\midrule

\multirow{2}{*}{\parbox{3cm}{Character judgement}}
& \textbf{People like YOU} are the reason so many people think they [...] \\
& Yeah \textbf{you sound like} a crazy person. \\
\midrule

\multirow{2}{*}{\parbox{3cm}{Confrontational questioning}}
& I'm not \textbf{sure why you} feel the need to put words into my mouth. \\
& Why \textbf{do you think} that I’ve been susceptible to brainwashing, but not you? \\

\midrule
\midrule

\multirow{1}{*}{\parbox{3cm}{Epistemic softening}}
& \textbf{I would argue} this definition doesn't go nearly far enough [...]\\
& I get \textbf{that the point} of war games is to [...] \\

\midrule
\multirow{2}{*}{\parbox{3cm}{Disagreement without challenge}}
& \textbf{Would you say} the word "pen" has lost all meaning and is not useful? \\
& \textbf{Even if you} take this as 1 person, he can work on getting fit. \\

\midrule
\multirow{2}{*}{\parbox{3cm}{Meta-linguistic clarification}}
& when we come \textbf{to talk about} what "cheating" actually is [...] \\
& Op is referencing people who \textbf{use the word} community to mean [...] \\

\bottomrule

\end{tabular}
\caption{Paraphrased examples of prototypical post-trigger (top) and post-deferral (bottom) replies exhibiting some of the most distinguishing phrases (\textbf{bolded}) identified by \citet{monroe_fightin_2017}'s distinguishing-words method.
Additional distinguishing phrases and sentences are listed in \autoref{tab:fw_ngrams_30} and \autoref{tab:ngram_addtl_examples} in \autoref{appendix:appendixsection}.
}
\label{tab:nondelayed_exs}
\end{table*}

Having established that adding an explicit decision making component yields substantial reductions in false positive rates, we now turn to an analysis of the model's decisions.
Our goal is to examine when the model chooses to defer triggering and why, and to assess whether these decisions align with the intuitions that motivated our approach---namely, that some tense moments admit plausible paths to recovery and therefore warrant patience.
By inspecting representative cases, we aim to ground the quantitative gains in interpretable conversational dynamics.

\xhdr{Anticipating decrease in tension}
One of the core intuitions motivating our approach is that humans appear to anticipate when a spike in conversational tension is likely to subside, and selectively avoid triggering in such moments.
To assess whether our system exhibits a similar behavior, we examine cases in which the decision mechanism deferred triggering at moments where the SOTA model would have otherwise triggered an alert.
For these moments, we conduct a post-hoc analysis on the test data and compute the subsequent change in estimated tension, $\mathcal{D}_k$, finding that in 83.5\% of cases the deferral is indeed followed by a decrease in tension ($\mathcal{D}_k > 0$).
In contrast, the baseline probability of observing a decrease in tension following an arbitrary SOTA trigger is substantially lower (55.3\%), suggesting that our model’s deferrals are aligned with anticipations of recovery.

\xhdr{Qualitative analysis}
To better understand the types of recovery the model is able to anticipate, we conduct a qualitative content analysis of the replies that immediately follow deferred triggering decisions.
As a first step toward a systematic exploration, we perform a Bayesian distinguishing-words analysis \cite{monroe_fightin_2017},\footnote{We use $\tau=5$ in this analysis for broader coverage. Observations remain qualitatively similar for $\tau = 7$ (\autoref{appendix:appendixsection}).} comparing replies that follow utterances where the decision to trigger was deferred (henceforth, \textit{post-deferral replies}) with replies that follow utterances where the trigger was maintained (\textit{post-trigger replies}).\footnote{These are all cases in which the SOTA would have triggered.  We discard cases in which the reply is the last utterance in a conversation or the to-be-forecasted personal attack.}

The phrases distinguishing the post-trigger replies (summarized in \autoref{tab:nondelayed_exs} and listed in \autoref{tab:fw_ngrams_30} in \autoref{appendix:appendixsection})
appear to use escalating language \cite{habernal_before_2018,zhang_conversations_2018}.
This includes direct accusations or attributions of fault, showing heavy usage of second-person pronouns (``you don['t] even'', ``the fact that you'', ``you think that''), 
implicit or explicit character judgment (``people like you'', ``you sound like'', ``for someone who''), 
and confrontational questioning (``why don't you'', ``sure why you'', ``not sure why'').

On the other side, the phrases distinguishing the post-deferral replies show signs of face-saving language \cite{brown_politeness_1987}  and repair attempts \cite{drew_open_1997,tsai_leveraging_2024}.
Some apparent strategies include epistemic softening and argument framing to distance the disagreement from the interlocutor (``would argue that'',``that the point'', ``that they don't''), the use of hypotheticals which allows the exploration of disagreements without direct challenge (``even if you'', ``would you say''), and meta-linguistic clarifications which reopen the interpretation of prior utterances and suggest that disagreement may arise from ambiguity rather than hostility (``use the word'', ``of the word'', ``to talk about''). 

Taken together, this contrast between post-deferral and post-trigger replies suggests that the decision mechanism is able to effectively anticipate conversational recovery, and use it to selectively defer triggering an alert.

\section{Conclusion}
\label{sec:conclusions}
In this work, we argue for the importance of a distinct decision-making component in conversational forecasting systems.
We introduce one such mechanism, which defers triggering when it anticipates---via simulation---plausible paths to conversational recovery.
While simple, this approach demonstrates how explicitly reasoning about when not to act can substantially improve forecasting behavior without sacrificing accuracy.
More broadly, our findings highlight the value of treating decision making as a first-class component of forecasting systems. 
This opens the door to richer and more flexible triggering mechanisms beyond fixed thresholding, such as mechanisms incorporating information about the speaker or decision policies inferred from the training data via reinforcement learning. 

From a deployment perspective, these results have direct implications for the design of real-world moderation and conversational support systems.
In high-stakes settings, such systems must balance the risk of acting too late against that of acting unnecessarily, as spurious alerts can erode user trust, overwhelm moderators, and disrupt otherwise recoverable conversations \cite{schluger_proactive_2022,chang_thread_2022}.
Our decision-making mechanism offers a principled way to manage this tradeoff by explicitly reasoning about deferral, with the parameter $\tau$ providing a direct and interpretable handle for end users to control how much evidence of plausible conversational recovery is required before intervening.
By enabling systems to anticipate and respect moments of potential de-escalation, such mechanisms enable more selective, context-sensitive conversational support interventions.

\section{Limitations}
\label{sec:limitations}
\xhdr{Scope of the decision mechanism}
In this work, we introduce a single, deliberately simple decision-making mechanism to demonstrate the value of decoupling belief estimation from triggering decisions.
While effective, this mechanism represents only one point in a much broader design space.
In particular, the exploration-exploitation tradeoff inherent in the unknown-horizon forecasting setting naturally aligns with reinforcement learning formulations, where a decision policy could be learned directly from training data rather than specified procedurally.
Exploring such learned policies---potentially optimizing downstream objectives such as false-positive costs or user experience---remains an important direction for future work.

Our decision mechanism focuses exclusively on deferring triggering decisions in tense moments when recovery appears plausible.
{This naturally introduces a precision-recall tradeoff where higher deferral may increase precision at the cost of recall.}
However, deferral is not the only possible intervention.
For example, another mechanism could instead \emph{hasten} triggering when simulated continuations suggest rapid escalation, even if the current estimated probability has not yet crossed the threshold.
More broadly, future systems could dynamically adjust when to act in either direction (defer or hasten), enabling richer and more flexible control over forecasting behavior.

\xhdr{Simulation fidelity and efficiency}
Our approach relies on large language model simulations to approximate likely next utterances.
While such simulations are increasingly strong, they are necessarily imperfect.
Moreover, our current implementation simulates only a single conversational step ahead, which limits its ability to capture longer-range recovery or escalation trajectories. 
This is due to the exponential increase in the number of simulations necessary, as each simulation would initiate its own branching of $k$ simulations.
Simulating further steps ahead, while costly due to the significant computation necessary for multiple LLM calls, may improve performance.
Future work could explore simulating full conversational rollouts, as well as methods for aggregating uncertainty across longer-term simulated futures.

\xhdr{Human baseline scale and coverage}
We use a human experiment to gain insight into how people make online forecasting decisions, prioritizing a small but carefully controlled and high-quality dataset.
As a result, the human baseline is limited in scale.
Future work could use our experimental methodology (which we open-source) to extend this baseline to a larger and more diverse set of conversations, include additional participant populations, or examine learning effects across more rounds to understand how---and to what extent---humans improve with experience.

\xhdr{Task, domain, and language specificity}
All experiments and analyses in this work are conducted within the context of the Conversations Gone Awry (CGA) task, which focuses on online, English-language discussions in a small number of communities.
As a result, both the belief estimation models and the proposed decision mechanism may reflect domain-, culture-, and language-specific norms tied to escalation.
While the methodology itself is not inherently tied to CGA, extending it to other domains would require careful adaptation and validation.
In principle, similar decision mechanisms could be applied to other forms of conversational forecasting, such as anticipating disengagement or breakdown in medical, educational, or support-oriented interactions, where the nature of both escalation and recovery may differ substantially.
We advocate for the development of such diverse conversational forecasting benchmarks.

\section*{Ethical Considerations} 

\noindent\textbf{Risk of inappropriate intervention.}
Applications of conversational derailment forecasting might involve intervening in ongoing human interactions.
While our decision mechanism reduces false positives, any conversational support system still carries the risk of intervening when intervention is unnecessary or unwanted.
Such interventions may disrupt legitimate disagreement, suppress minority viewpoints, or be perceived as intrusive.
Careful calibration of triggering behavior---including explicit control over parameters such as $\tau$---is therefore essential when deploying such systems in real-world conversational support settings. 
Careful user studies are necessary before real-world deployment and should consider the effects of the decision making mechanism when deployed in conversational support systems.

\noindent\textbf{Reliance on simulated futures.}
Our approach uses large language model simulations to anticipate possible conversational trajectories.
These simulations reflect the biases, limitations, and normative assumptions of the underlying language models, and may not faithfully represent how a conversation would actually evolve.
As a result, the decision mechanism may defer or trigger based on simulated recoveries or escalations that would not occur in practice.
This limitation underscores the importance of treating simulations as advisory signals rather than ground truth, and of validating decision mechanisms in deployment-specific contexts.

\noindent\textbf{Dataset and annotation biases.}
The CGA datasets reflect particular communities (e.g., Change My View and Wikipedia talk pages) and moderation norms.
Consequently, both the belief estimation models and the decision mechanisms trained or evaluated on these datasets may encode community-specific notions of what constitutes ``tension'' or ``derailment.''
Applying such systems to other conversational domains without careful adaptation risks mischaracterizing culturally or contextually appropriate forms of disagreement.

\noindent\textbf{Human subject considerations.}
Our human baseline experiment involves participants making judgments about potentially contentious or hostile conversational content.
All participants were informed their annotations would be used to create a public dataset.
The task was designed to minimize exposure to explicit personal attacks (the attacks were not displayed) and the experiment was vetted by an institutional IRB.
Repeated exposure to tense interactions may still impose cognitive or emotional burden.
Future large-scale studies should consider additional safeguards, including clear opt-out mechanisms and post-task debriefing, especially if extended to more adversarial domains.

\xhdr{Additional considerations}
The experiments in this work use datasets containing real-world conversations from the r/ChangeMyView subreddit and Wikipedia Talk Pages. In examples of paraphrased comments from users, we anonymize speakers. Throughout this work, the data used is publicly accessible and we use only open-source models in our experiments.

\section{Acknowledgments}
\label{sec:acknowledgements}
Foremost, we offer our \textit{deferred} thanks to Jonathan P. Chang for very early discussions about ``ways out'' (which he eventually found!) and to Nicholas Chernogor, Tushaar Gangavarapu, and Son Tran for developing the infrastructure on which this work builds.
We are grateful for engaging discussions with the other members of the Team Zissou---including Dave Jung, Lillian Lee, Ethan Xia, and Sean Zhang.
We thank the reviewers for their extremely thoughtful feedback.
We gratefully acknowledge the use of research computing resources from the Empire AI Consortium, Inc., supported by Empire State Development of the State of New York, the Simons Foundation, and the Secunda Family Foundation.  
This work was in part enabled by a Gemma Academic Program GCP Credit Award.
This project was developed as part of Cornell's Bowers Undergraduate Research Experience (BURE) program in which Laerdon Kim participated.
Vivian Nguyen was supported by a Cornell University Computer Science Fellowship and a Cornell Graduate School Dean’s Scholarship.
Cristian Danescu-Niculescu-Mizil was funded in part by the U.S. National Science Foundation under Grant No. IIS-1750615 (CAREER), by Cornell’s Center for Social Sciences, by a LinkedIn Research Award, and by a Wikimedia Research Fund Award.
Any opinions, findings, and conclusions in this work are those of the author(s) and do not necessarily reflect the views of Cornell University or the National Science Foundation.

\bibliography{refs}

@inproceedings{nguyen_hanging_2025,
	author = {Vivian Nguyen and Lillian Lee and Cristian Danescu-Niculescu-Mizil},
	year = {2025},
	title = {Hanging in the {Balance}: {Pivotal} {Moments} in {Crisis} {Counseling} {Conversations}},
	booktitle = {Proceedings of {ACL}},
}

@inproceedings{zhang_conversations_2018,
	author = {Justine Zhang and Jonathan Chang and Cristian Danescu-Niculescu-Mizil and Lucas Dixon and Yiqing Hua and Dario Taraborelli and Nithum Thain},
	year = {2018},
	title = {Conversations {Gone} {Awry}: {Detecting} {Early} {Signs} of {Conversational} {Failure}},
	booktitle = {Proceedings of {ACL}},
}

@inproceedings{chang_trouble_2019,
	author = {Jonathan P. Chang and Cristian Danescu-Niculescu-Mizil},
	year = {2019},
	title = {Trouble on the {Horizon}: {Forecasting} the {Derailment} of {Online} {Conversations} as {They} {Develop}},
	booktitle = {Proceedings of {EMNLP-IJCNLP}},
}

@phdthesis{chang2024thesis,
    author = {Jonathan P. Chang},
    title = {Towards {Computational} {Methods} for {Proactively} {Supporting} {Healthier} {Online} {Discussions}},
    school = {Cornell University},
    year = {2024},
    url = {https://www.cs.cornell.edu/~cristian/papers/chang_thesis.pdf},
}

@inproceedings{olpadkar_can_2025,
	author = {Kaustubh Olpadkar and Vikram Sunil Bajaj and Leslie Barrett},
	year = {2025},
	title = {Can {LLMs} Be {Efficient} {Predictors} of {Conversational} {Derailment}?},
	booktitle = {Findings of {EMNLP}},
}

@article{monroe_fightin_2017,
	author = {Burt L. Monroe and Michael P. Colaresi and Kevin M. Quinn},
	year = {2017},
	title = {Fightin' {Words}: {Lexical} {Feature} {Selection} and {Evaluation} for {Identifying} the {Content} of {Political} {Conflict}},
	journal = {Political Analysis},
    volume = {16},
    number = {4},
    pages = {372--403},
}

@inproceedings{habernal_before_2018,
	author = {Ivan Habernal and Henning Wachsmuth and Iryna Gurevych and Benno Stein},
	year = {2018},
	title = {Before {Name-Calling}: {Dynamics} and {Triggers} of {Ad} {Hominem} {Fallacies} in {Web} {Argumentation}},
	booktitle = {Proceedings of {NAACL}},
}

@inproceedings{jurgens_just_2019,
	author = {David Jurgens and Libby Hemphill and Eshwar Chandrasekharan},
	year = {2019},
	title = {A {Just} and {Comprehensive} {Strategy} for {Using} {NLP} to {Address} {Online} {Abuse}},
	booktitle = {Proceedings of {ACL}},
}

@inproceedings{loshchilov_decoupled_2019,
	author = {Ilya Loshchilov and Frank Hutter},
	year = {2019},
	title = {Decoupled {Weight} {Decay} {Regularization}},
	booktitle = {Proceedings of {ICLR}},
}

@article{chang_thread_2022,
	author = {Jonathan P. Chang and Charlotte Schluger and Cristian Danescu-Niculescu-Mizil},
	year = {2022},
	title = {Thread {With} {Caution}: {Proactively} {Helping} {Users} {Assess} and {Deescalate} {Tension} in {Their} {Online} {Discussions}},
	journal = {Proceedings of {CSCW}},
    volume = {6},
    number = {545},
    pages = {1--37},
}

@misc{tran_conversations_2025,
	author = {Son Quoc Tran and Tushaar Gangavarapu and Nicholas Chernogor and Jonathan P. Chang and Cristian Danescu-Niculescu-Mizil},
	year = {2025},
	title = {Conversations {Gone} {Awry}, {But} {Then}? {Evaluating} {Conversational} {Forecasting} {Models}},
	archivePrefix = {arXiv},
	eprint = {2507.19470},
	primaryClass = {cs.CL},
    note = {arXiv:2507.19470},
}

@inproceedings{bao_conversations_2021,
	author = {Jiajun Bao and Junjie Wu and Yiming Zhang and Eshwar Chandrasekharan and David Jurgens},
	year = {2021},
	title = {Conversations {Gone} {Alright}: {Quantifying} and {Predicting} {Prosocial} {Outcomes} in {Online} {Conversations}},
	booktitle = {Proceedings of {WWW}},
}

@inproceedings{altarawneh_conversation_2023,
	author = {Enas Altarawneh and Ameeta Agrawal and Michael Jenkin and Manos Papagelis},
	year = {2023},
	title = {Conversation {Derailment} {Forecasting} with {Graph} {Convolutional} {Networks}},
	booktitle = {Proceedings of {WOAH}, co-located with {ACL}},
}

@inproceedings{yuan_conversation_2023,
	author = {Jiaqing Yuan and Munindar P. Singh},
	year = {2023},
	title = {Conversation {Modeling} to {Predict} {Derailment}},
	booktitle = {Proceedings of {ICWSM}},
}

@inproceedings{vannella_validating_2014,
	author = {Daniele Vannella and David Jurgens and Daniele Scarfini and Domenico Toscani and Roberto Navigli},
	year = {2014},
	title = {Validating and {Extending} {Semantic} {Knowledge} {Bases} using {Video} {Games} with a {Purpose}},
	booktitle = {Proceedings of {ACL}},
}

@inproceedings{korre_evaluation_2025,
	author = {Katerina Korre and Dimitris Tsirmpas and Nikos Gkoumas and Emma Cabalé and Danai Myrtzani and Theodoros Evgeniou and Ion Androutsopoulos and John Pavlopoulos},
	year = {2025},
	title = {Evaluation and {Facilitation} of {Online} {Discussions} in the {LLM} {Era}: {A} {Survey}},
	booktitle = {Proceedings of {EMNLP}},
}

@misc{li_using_2022,
	author = {Cathleen Li},
	year = {2022},
    month = {5},
    day = {3},
	title = {Using {Predictive} {Technology} to {Foster} {Constructive} {Conversations}},
	journal = {Medium},
    note = {Nextdoor Engineering Blog, May 3 2022. Retrieved April 19, 2026},
    url = {https://engblog.nextdoor.com/using-predictive-technology-to-foster-constructive-conversations-4af437942bd4},
    urldate = {2026-04-19},
}

@inproceedings{kementchedjhieva_dynamic_2021,
	author = {Yova Kementchedjhieva and Anders Søgaard},
	year = {2021},
	title = {Dynamic {Forecasting} of {Conversation} {Derailment}},
	booktitle = {Proceedings of {EMNLP}},
}

@article{schluger_proactive_2022,
	author = {Charlotte Schluger and Jonathan P. Chang and Cristian Danescu-Niculescu-Mizil and Karen Levy},
	year = {2022},
	title = {Proactive {Moderation} of {Online} {Discussions}: {Existing} {Practices} and the {Potential} for {Algorithmic} {Support}},
	journal = {Proceedings of {CSCW}},
    article = {370},
}

@inproceedings{zhang_forecasting_2025,
	author = {Yunfan Zhang and Kathleen McKeown and Smaranda Muresan},
	year = {2025},
	title = {Forecasting {Conversation} {Derailments} {Through} {Generation}},
	booktitle = {Proceedings of {INLG}},
}

@misc{teamGemma2Improving2024,
  author = {Gemma Team},
  year = {2024},
  title = {{Gemma} 2: {Improving} Open Language Models at a Practical Size},
  archivePrefix = {arXiv},
  eprint = {2408.00118},
  note = {arXiv:2408.00118 [cs.CL]}
}

@misc{grattafioriLlama3Herd2024,
  author = {Aaron Grattafiori and Abhimanyu Dubey and others},
  year = {2024},
  title = {The {Llama} 3 {Herd} of {Models}},
  archivePrefix = {arXiv},
  eprint = {2407.21783},
  primaryClass = {cs.CL},
  note = {arXiv:2407.21783}
}

@inproceedings{hu_lora_2022,
	author = {Edward J. Hu and Yelong Shen and Phillip Wallis and Zeyuan Allen-Zhu and Yuanzhi Li and Shean Wang and Lu Wang and Weizhu Chen},
	year = {2022},
	title = {{LoRA}: {Low-Rank} {Adaptation} of {Large} {Language} {Models}},
	booktitle = {Proceedings of {ICLR}},
}

@inproceedings{tsai_leveraging_2024,
	author = {Che Wei Tsai and Yen-Hao Huang and Tsu-Keng Liao and Didier Fernando Salazar Estrada and Retnani Latifah and Yi-Shin Chen},
	year = {2024},
	title = {Leveraging {Conflicts} in {Social} {Media} {Posts}: {Unintended} {Offense} {Dataset}},
	booktitle = {Proceedings of {EMNLP}},
}

@book{brown_politeness_1987,
	author = {Penelope Brown and Stephen C. Levinson},
	year = {1987},
	title = {Politeness: {Some} {Universals} in {Language} {Usage}},
	publisher = {Cambridge University Press},
}

@article{drew_open_1997,
	author = {Paul Drew},
	year = {1997},
	title = {`{Open}' {Class} {Repair} {Initiators} in {Response} to {Sequential} {Sources} of {Troubles} in {Conversation}},
	journal = {Journal of Pragmatics},
    volume = {28},
    number = {1},
}

\clearpage

\appendix
\section{Additional Examples}
\label{appendix:appendixsection}

\definecolor{lightblue}{RGB}{230,240,255}
\definecolor{mediumblue}{RGB}{150,160,200}

\begin{table*}[h]
\centering
\small
\setlength{\tabcolsep}{5pt}
\begin{tabular}{p{0.02\linewidth} p{0.10\linewidth} p{0.55\linewidth} p{0.12\linewidth} p{0.06\linewidth}}
\hline
\textbf{t} & \textbf{Author} & \textbf{Utterance (abridged)} & \textbf{\# of calm simulations} & \textbf{$\mathcal{P}$} \\
\hline
0 & Speaker 1 &
Has anyone ever actually tried to police DiCaprio's dating life? [...] &
10.0 & 0.50 \\

1 & Speaker 2 &
Declaring something as ``creepy'' [...] is essentially saying it's unacceptable behaviour. [...] ``shifting frames.'' &
0.0 & 0.59 \\

\rowcolor{lightblue}
2 & Speaker 3 &
So what is your opinion? That people shouldn't be allowed to call DiCaprio creepy because that's policing him? &
9.0 & \textbf{0.65} \\

3 & Speaker 2 &
The post is about those who argue against large age-gap relationships [...] So far, you're proving my point. &
7.0 & 0.50 \\

4 & Speaker 3 &
Your opinion is unclear to me because it doesn't sound like you want your
views changed regarding the arguments people make against age-gap
relationships, but rather you want your views on how you feel about the way
people are expressing said arguments/the effect that has changed. [...] I'm asking about conclusions you
draw from your position [...] I
think that's a fair question to ask and I'm not engaging in bad faith here. &
9.0 & 0.59 \\
\hline
\end{tabular}
\caption{Example thread from CGA-CMV-Large where decision-deferral is effective; conversation does not have a removed comment, and thus remains calm. $\mathcal{P} := \pderail$; $\mathcal{P}$ values above $T$ are highlighted in \textbf{bold}, which are moments where SOTA will trigger. An utterance which has a trigger decision deferral is highlighted in \textcolor{mediumblue}{blue}.}
\label{tab:conv_example_win1}
\end{table*}

\begin{table*}[h]
\centering
\small
\setlength{\tabcolsep}{5pt}
\begin{tabular}{p{0.02\linewidth} p{0.10\linewidth} p{0.55\linewidth} p{0.12\linewidth} p{0.06\linewidth}}
\hline
\textbf{t} & \textbf{Author} & \textbf{Utterance (abridged)} & \textbf{\# of calm simulations} & \textbf{$\mathcal{P}$} \\
\hline
0 & Speaker 1 &
You're applying human logic and motivations to an omniscient deity[...] humanity is fundamentally incapable of understanding the Christian God[...] we're like 2 year olds. &
10.0 & 0.35 \\

1 & Speaker 2 &
That sounds like some shit Rian Johnson would say [...] God is a primitive attempt at a Thanos-style big baddie [...] inconsistent due to many writers [...] &
1.0 & 0.59 \\

\rowcolor{lightblue}
2 & Speaker 3 &
I don't see how this contributes [...] You can't just discount someone's comment [...] You need to actually address what they said with counterpoints. &
9.0 & \textbf{0.65} \\

3 & Speaker 2 &
> But humanity is fundamentally incapable [...] However we’re perfectly capable of understanding a fictional character [...] inconsistencies are expected from many writers [...] There, that better? &
9.0 & 0.47 \\

4 & Speaker 3 &
No because all you have done is paraphrased the title of the CMV. &
8.0 & 0.53 \\ \\

& & [Removed by moderator due to Rule 2 violation] \\\\

\hline
\end{tabular}
\caption{Example thread from CGA-CMV-Large where decision-deferral is not effective; conversation has a removed comment, and thus goes awry.}
\label{tab:conv_example_loss1}

\vspace{0.75cm}

\centering
\small
\setlength{\tabcolsep}{6pt}
\begin{tabular}{ c p{0.65\textwidth}}
\toprule
\textbf{$\mathcal{P}$} & \textbf{Simulated next utterance (abridged)} \\
\midrule
0.53 &
The point of my post is literally about why I think your view is wrong [...] which again means it did contribute to the discussion. \\

0.50 &
This response is so weirdly confrontational [...] My point wasn't to attack anyone personally, but rather show why the premise itself has flaws [...] \\

0.50 &
How does mine not? \\

0.47 &
He didn’t create humans to torture them though [...] That idea is based on a very specific interpretation [...] The rest of the New Testament doesn’t back up your view here. \\

0.50 &
Because OP asked about cognitive dissonance within Christianity [...] those conflicts exist because the Bible was written by multiple people over many years. \\

\textbf{0.65} &
This guy seems like one of those people who thinks there’s something wrong with him [...] This person is simply unreasonable. \\

0.50 &
Of course it’s ridiculous when seen outside its context [...] it comes from people trying to give form to things bigger than themselves. \\

0.44 &
So there’s no challenge? The fun of creation comes from taking nothing and building something out of it [...] \\

0.50 &
The concept of Satan comes from Zoroastrianism [...] Hell does not exist until the New Testament [...] \\

0.38 &
Since God cannot be known, we should not concern ourselves with him [...] focus instead on human suffering and ethics. \\
\bottomrule
\end{tabular}
\caption{
Simulated next utterances and their forecast probabilities at $t=2$, predicting comments for $t=3$.
}
\label{tab:sim_reply_example}
\end{table*}

\xhdr{Examples of decision-deferral on CGA-CMV-Large}
To concretize the observations made via our qualitative analysis, we show specific examples of successful and unsuccessful trigger decision deferrals.
In \autoref{tab:conv_example_win1}, at $t=2$, we notice the tension in the conversation beginning to increase, where Speaker 3 seems to question the opinions of the previous speakers.
However, because 9/10 of our simulations for the next utterance indicate a prediction that the conversation  will not go awry, we choose not to make our awry prediction yet.
At $t=4$, we see that the tension in the conversation has decreased, clearing up that their question at $t=2$ was perhaps a genuine question meant to clarify the disagreement, rather than to antagonize the other discussion participants.
Without the deferral, the forecaster would have triggered intervention, stymieing this path to recovery.

In our manual analysis of instances where the deferral helped the forecaster avoid a false positive error, we find that many instances fall into this pattern, of an inaccurate forecast being made due to a seemingly awry utterance which ends up being a misinterpretation, or eventually recovering.

\autoref{tab:conv_example_loss1} illustrates an example where decision-deferral fails to recognize a derailment, whereas the SOTA does.
The utterance at $t=1$ introduces some tension, with Speaker 2 criticizing the comment of Speaker 1 with an expletive.
Speaker 3 joins the conversation to claim that Speaker 2 is not productively contributing to the discussion, which is the first utterance where an awry forecast is made, and deferred, as the utterance has 9/10 calm simulations.
After this comment, Speaker 2 slightly adjusts their tone, ending with a sarcastic comment and an open ended question.
We notice that $D_2 > 0$ in this example, but the conversation appears to re-escalate after Speaker 3 shuts down Speaker 2 again, thus increasing the inferred tension.
Here, the decision-deferral system arguably did correctly predict that the next utterance would indeed be a de-escalation with an inferred tension below $T$.
Afterwards, the conversation re-escalated, but not to the point where the forecaster could make an intervention.
This example demonstrates the complexity of handling decision-making in time-series; it is ambiguous as to whether the system ``should have'' intervened at $t=2$ or $t=4$.

Although there does not exist an awry forecast at $t=3$, we might understand the quote ``There, that better?'' to be more combative, whereas the forecaster might focus on the bulk of the comment being revised for tone.
It is perhaps due to this underestimation of Speaker 2's sarcasm, and Speaker 3's subsequent contradiction of Speaker 2, that the conversation unexpectedly goes awry.
In addition, none of the simulated utterances predict the sarcasm which follows the utterance at $t=2$.

\xhdr{N-gram phrases and in-context examples}
We provide additional examples of n-gram phrases (\autoref{tab:fw_ngrams_30}, \autoref{tab:fw_ngrams_tauX}) and examples with conversational context  (\autoref{tab:ngram_addtl_examples}, \autoref{tab:ngram_addtl_examples_tau7}). We run distinguishing words analysis with $n=3$.

\begin{table*}[h]
\centering
\small
\setlength{\tabcolsep}{6pt}
\begin{tabular}{p{0.14\textwidth} r | p{0.14\textwidth} r}
\toprule
\multicolumn{2}{c|}{\textbf{Not deferred n-grams}} &
\multicolumn{2}{c}{\textbf{Deferred n-grams}} \\
\cmidrule(lr){1-2} \cmidrule(lr){3-4}
\textbf{n-gram} & \textbf{$z$} &
\textbf{n-gram} & \textbf{$z$} \\
\midrule
fact that you        & $-3.58$ & they are the        & $+3.39$ \\
on the internet      & $-2.97$ & would argue that   & $+3.21$ \\
you know it          & $-2.75$ & because they are   & $+3.02$ \\
people who are       & $-2.65$ & have nothing to    & $+2.90$ \\
you don even         & $-2.65$ & to support the     & $+2.74$ \\
you think that       & $-2.62$ & is no reason       & $+2.74$ \\
all you want         & $-2.58$ & if you re          & $+2.70$ \\
people like you      & $-2.55$ & right to life      & $+2.67$ \\
that you re          & $-2.45$ & to talk about      & $+2.54$ \\
why don you          & $-2.44$ & know for fact      & $+2.52$ \\
sure why you         & $-2.43$ & of the word        & $+2.50$ \\
and not the          & $-2.43$ & of course there    & $+2.49$ \\
in the west          & $-2.43$ & the only way       & $+2.47$ \\
you have any         & $-2.37$ & use the word       & $+2.47$ \\
you want but         & $-2.36$ & because it is     & $+2.46$ \\
to back up           & $-2.33$ & that makes you    & $+2.42$ \\
the age of           & $-2.33$ & to make it        & $+2.36$ \\
that you ve          & $-2.33$ & the ability to    & $+2.34$ \\
of the best          & $-2.29$ & it doesn matter   & $+2.33$ \\
what you are         & $-2.23$ & would you say     & $+2.33$ \\
that would be        & $-2.22$ & then you can      & $+2.31$ \\
to go back           & $-2.22$ & it was in         & $+2.30$ \\
not sure why         & $-2.22$ & that the point    & $+2.26$ \\
just look at         & $-2.20$ & they should have  & $+2.26$ \\
you need to          & $-2.20$ & you re not        & $+2.26$ \\
for someone who      & $-2.20$ & in front of       & $+2.24$ \\
if they had          & $-2.20$ & even if you       & $+2.24$ \\
to me that           & $-2.20$ & that they don     & $+2.24$ \\
you sound like       & $-2.18$ & only way to       & $+2.21$ \\
you understand that  & $-2.18$ & talk about it     & $+2.21$ \\
\bottomrule
\end{tabular}
\caption{Top 30 tri-grams associated with replies following not-deferred (left) and deferred (right) decisions, ranked by the z-score of the Bayesian distinguishing-word analysis for $\tau = 5$.}
\label{tab:fw_ngrams_30}
\end{table*}

\begin{table*}[h]
\centering
\small
\setlength{\tabcolsep}{6pt}
\begin{tabular}{p{0.14\textwidth} r | p{0.14\textwidth} r}
\toprule
\multicolumn{2}{c|}{\textbf{Not deferred n-grams}} &
\multicolumn{2}{c}{\textbf{Deferred n-grams}} \\
\cmidrule(lr){1-2} \cmidrule(lr){3-4}
\textbf{n-gram} & \textbf{$z$} &
\textbf{n-gram} & \textbf{$z$} \\
\midrule
fact that you        & $-4.46$ & to talk about     & $+2.92$ \\
people who are       & $-3.50$ & because it is     & $+2.80$ \\
you know it          & $-3.03$ & in front of       & $+2.77$ \\
because that is      & $-2.79$ & they are the      & $+2.70$ \\
do you see           & $-2.78$ & if you re         & $+2.68$ \\
if you go            & $-2.78$ & would argue that  & $+2.68$ \\
of the best          & $-2.66$ & have nothing to   & $+2.60$ \\
sure why you         & $-2.58$ & but they are      & $+2.52$ \\
people like you      & $-2.54$ & you can have      & $+2.50$ \\
the age of           & $-2.54$ & right to life     & $+2.49$ \\
the guy who          & $-2.54$ & because they are  & $+2.42$ \\
of that is           & $-2.53$ & it doesn matter   & $+2.41$ \\
not just the         & $-2.53$ & the only way      & $+2.31$ \\
think you know       & $-2.53$ & that women are    & $+2.28$ \\
that you re          & $-2.48$ & but that doesn    & $+2.24$ \\
you talking about    & $-2.47$ & if they re        & $+2.20$ \\
if you would         & $-2.45$ & more or less      & $+2.17$ \\
needs to be          & $-2.44$ & there is more     & $+2.17$ \\
because you don      & $-2.41$ & it is in          & $+2.17$ \\
what you are         & $-2.39$ & should not be     & $+2.16$ \\
not sure why         & $-2.38$ & talk about it     & $+2.16$ \\
what wrong with      & $-2.38$ & the us but        & $+2.14$ \\
say that you         & $-2.38$ & use the word      & $+2.14$ \\
example of how       & $-2.38$ & having sex with   & $+2.14$ \\
right to be          & $-2.36$ & is no reason      & $+2.14$ \\
why don you          & $-2.35$ & to support the    & $+2.14$ \\
to have any          & $-2.33$ & the will of       & $+2.14$ \\
to stop the          & $-2.33$ & but if you        & $+2.14$ \\
tell me how          & $-2.31$ & what you re       & $+2.12$ \\
you don even         & $-2.31$ & what they are     & $+2.10$ \\
\bottomrule
\end{tabular}
\caption{Top 30 tri-grams associated with replies following not-deferred (left) and deferred (right) decisions, ranked by the z-score of the Bayesian distinguishing-word analysis for $\tau = 7$.}
\label{tab:fw_ngrams_tauX}
\end{table*}

\begin{table*}[t]
\centering
\begin{tabular}{p{3cm} p{12cm}}
\toprule
\textbf{Phenomenon} & \textbf{Example (distinguishing phrase in bold)} \\
\midrule

\multirow{1}{*}{\parbox{3cm}{Confrontational questioning}}
& >> Yeah I don’t think this is effective. [...]
Also taxing assets the same as income is impossible because assets aren’t liquid. That means you can be getting taxed either too much, or too little for the cash equivalent of the assets you receive. That’s why they’re on an entirely separate tax bracket than regular income. I think you need to sit down and study taxes instead of Econ tbh 

>> Right which is why I specified the value on the day at which the the stock was issued. It's not a perfect system, but it's better than having completely uncontrolled CEO pays is my point. Yes I understand that stocks can vary in value, which is why I'm not \textbf{sure why you}'re talking about taxing them. I never said anything about taxes? 
\\
\midrule

\multirow{2}{*}{\parbox{3cm}{Character judgement}}
& >> I'm a teacher and this is leading reason I will never ever return to America. I hate school shootings, I hate active shooter drills, I hate militarized schools designed as high security facilities. And I hate that pro-gun \textbf{people like you} refuse every single change proposed, but then propose no other changes. \\

\midrule
\midrule

\multirow{1}{*}{\parbox{3cm}{Epistemic softening}}
& >> No one denies that having nukes is good for the north korean regime.   what people deny is the idea that the most monstrous regime on the planet has any legitimacy or that its desires deserve to be respected.

>> Define monstrous, \textbf{I would argue} that our hawkishness has led to far more bloodshed than anything North Korea has done. Very patriocentric of you. I would agree they are the most monstrous regime to their citizens, but not far ahead of Saudi Arabia. 
\\

\bottomrule

\end{tabular}
\caption{Additional paraphrased examples with conversational context of prototypical post-trigger (top) and post-deferral (bottom) replies exhibiting some of the most distinguishing phrases (\textbf{bolded}) identified by \citet{monroe_fightin_2017}'s distinguishing-words method $(\tau = 5)$.
}
\label{tab:ngram_addtl_examples}
\end{table*}

\begin{table*}[t]
\centering
\begin{tabular}{p{3cm} p{12cm}}
\toprule
\textbf{Phenomenon} & \textbf{Example (distinguishing phrase in bold)} \\
\midrule

\multirow{1}{*}{\parbox{3cm}{Direct attribution of fault}}
& It's apparent you believe \textbf{you know it} all and no one is ever going to change your mind.  \\
\midrule

\multirow{2}{*}{\parbox{3cm}{Character judgement}}

& Basically, \textbf{you sound like} a pretentious art snob \\
& I don't \textbf{think YOU know} what "full reserve banking" is... \\
\midrule
\multirow{2}{*}{\parbox{3cm}{Confrontational questioning}}
& \textbf{Tell me how} Jan 6 does not fit the parameters of the textbook definition of insurrection? \\

\midrule
\midrule

\multirow{1}{*}{\parbox{3cm}{Epistemic softening}}
& I \textbf{would argue that} in most cases, ie the woman having consensual sex, that the creation of life is what entitles the fetus/child. \\

\midrule
\multirow{2}{*}{\parbox{3cm}{Disagreement without challenge}}
& \textbf{Would you say} that someone who voted for segregationists during the civil rights era was not necessarily a racist? \\

\midrule
\multirow{2}{*}{\parbox{3cm}{Meta-linguistic clarification}}
& ...when we come \textbf{to talk about} what "cheating" actually is... \\
& ...so paedophiles do not tend to \textbf{talk about it} openly... \\

\bottomrule

\end{tabular}
\caption{Additional paraphrased examples of prototypical post-trigger (top) and post-deferral (bottom) replies exhibiting some of the most distinguishing phrases (\textbf{bolded}) identified by \citet{monroe_fightin_2017}'s distinguishing-words method $(\tau = 7)$.
}
\label{tab:ngram_addtl_examples_tau7}
\end{table*}

\section{Additional Baselines and Ablations}
\label{appendix:morebaselines}
\begin{table*}[t]
\newcommand{\sd}[1]{{\scriptsize$\pm$#1}}
\centering
{
\begin{tabular}{lcccccc}
\toprule
Method & Acc $\uparrow$ & FPR $\downarrow$ & P $\uparrow$ & R $\uparrow$ & F1 $\uparrow$ & {H} \\
\midrule
SOTA         & \textbf{70.9} \sd{0.35} & 34.3 \sd{3.96} & 69.1 \sd{1.21} & 76.1 \sd{4.33} & \textbf{72.3} \sd{1.30} & 3.9 \sd{0.13} \\
\textbf{+ selective deferral}  & \textbf{70.9} \sd{0.77} & \textbf{26.7} \sd{3.10} & \textbf{72.1} \sd{0.92} & 68.4 \sd{4.63} & 70.1 \sd{1.96} & 3.8 \sd{0.09} \\
 + random deferral (train tuned) & 69.4 \sd{0.79} & 30.2 \sd{3.81} & 70.0 \sd{1.30} & 69.0 \sd{0.48} & 69.2 \sd{1.91} & 3.8 \sd{0.12} \\
 + random deferral (validation tuned) & 69.0 \sd{0.98} & 29.3 \sd{3.80} & 69.8 \sd{1.44} & 67.1 \sd{4.83} & 68.3 \sd{2.10} & 3.8 \sd{0.12} \\
   + random deferral (test tuned) & 70.2 \sd{0.78} & 32.0 \sd{3.88} & 69.4 \sd{1.33} & 72.3 \sd{4.57} & 70.8 \sd{1.69} & 3.9 \sd{0.12} \\
\midrule
+ simulation (average)    & 70.2 \sd{0.27} & 36.2 \sd{5.17} & 68.1 \sd{1.52} & \textbf{76.6} \sd{5.28} & 72.0 \sd{1.45} & 4.0 \sd{0.11} \\
+ simulation (majority) & 70.0 \sd{0.29} & 36.9 \sd{5.12} & 67.7 \sd{1.57} & 76.7 \sd{4.81} & 71.8 \sd{1.21} & 4.0 \sd{0.11} \\
\midrule
\midrule
+ variance deferral & 70.9 & 33.2 & 69.6 & 75.0 & 72.2 & 3.8 \\
\midrule
\midrule
oracle threshold ($\tau = 7$)  & 70.0 \sd{0.88} & 26.7 \sd{3.31} & 71.5 \sd{1.00} & 66.8 \sd{4.94} & 69.0 \sd{2.18} & 3.7 \sd{0.10}\\
\bottomrule
\end{tabular}
}
\caption{Average performance on CGA-CMV-Large across 5 random seeds. For variance deferral, no averaging is possible, as the method requires five separate seeds. Standard deviation is reported to the right of each table entry.}.
\label{tab:cmv_avg_perf_appendix}
\end{table*}

\begin{table*}[h]
\centering
\normalsize
\begin{tabular}{clcccccr}
\toprule
$\tau$ & Method & Acc $\uparrow$ & FPR $\downarrow$ & P $\uparrow$ & R $\uparrow$ & F1 $\uparrow$ & H \\
\midrule
\multirow{2}{*}{1}
  & selective deferral  & 67.2 & 16.8 & 75.5 & 51.3 & 60.9 & 3.5 \\
  & matched FPR oracle  & 66.5 & 16.6 & 75.1 & 49.8 & 59.7 & 3.4 \\
\addlinespace[4pt]
\multirow{2}{*}{2}
  & selective deferral  & 68.2 & 19.0 & 74.7 & 55.4 & 63.5 & 3.6 \\
  & matched FPR oracle  & 67.6 & 18.8 & 74.3 & 53.9 & 62.4 & 3.5 \\
\addlinespace[4pt]
\multirow{2}{*}{3}
  & selective deferral  & 69.0 & 20.4 & 74.2 & 58.5 & 65.3 & 3.6 \\
  & matched FPR oracle  & 68.0 & 20.2 & 73.7 & 56.2 & 63.6 & 3.5 \\
\addlinespace[4pt]
\multirow{2}{*}{4}
  & selective deferral  & 69.7 & 21.7 & 73.9 & 61.1 & 66.8 & 3.6 \\
  & matched FPR oracle  & 68.8 & 22.2 & 73.0 & 59.9 & 65.7 & 3.6 \\
\addlinespace[4pt]
\multirow{2}{*}{5}
  & selective deferral  & 70.0 & 23.3 & 73.2 & 63.2 & 67.8 & 3.7 \\
  & matched FPR oracle  & 68.9 & 22.9 & 72.8 & 60.7 & 66.0 & 3.6 \\
\addlinespace[4pt]
\multirow{2}{*}{6}
  & selective deferral  & 70.6 & 24.8 & 72.8 & 65.9 & 69.1 & 3.7 \\
  & matched FPR oracle  & 69.6 & 25.3 & 72.0 & 64.4 & 67.8 & 3.7 \\
\addlinespace[4pt]
\multirow{2}{*}{7}
  & selective deferral  & 70.9 & 26.7 & 72.1 & 68.4 & 70.1 & 3.8 \\
  & matched FPR oracle  & 70.0 & 26.7 & 71.5 & 66.8 & 69.0 & 3.7 \\
\addlinespace[4pt]
\multirow{2}{*}{8}
  & selective deferral  & 71.1 & 28.9 & 71.2 & 71.2 & 71.1 & 3.8 \\
  & matched FPR oracle  & 70.5 & 29.7 & 70.5 & 70.6 & 70.5 & 3.8 \\
\addlinespace[4pt]
\multirow{2}{*}{9}
  & selective deferral  & 71.2 & 31.5 & 70.2 & 73.9 & 72.0 & 3.9 \\
  & matched FPR oracle  & 70.6 & 31.3 & 70.0 & 72.5 & 71.1 & 3.8 \\
\bottomrule
\end{tabular}
\caption{Per-$\tau$ comparison of selective deferral against a
matched FPR oracle baseline on CGA-CMV-Large (average over 5 seeds).
The oracle baseline sets the forecaster threshold to match the deferral FPR at each $\tau$.}
\label{table:taumatching}
\end{table*}

We supply additional baseline trials and report standard deviations across seeds in \autoref{tab:cmv_avg_perf_appendix}.

\xhdr{Variance deferral baseline}
We might also use variance between models fine-tuned using different random seeds to examine moments at which there is higher disagreement of tension estimation.
Using our five random seeds, we can again ablate the selective simulation component and instead defer based on the variance calculated between the five forecast probabilities using a threshold tuned for accuracy on the validation set.

\xhdr{Random deferral tuning}
Instead of finding our probability of deferral from the train set, we can also source this probability from the test set.

\xhdr{Deferral with variable tau and FPR-matched oracle}
In \autoref{table:taumatching}, we provide more detailed tau-matching using the same procedure with which oracle threshold in \autoref{tab:cmv_avg_perf} was calculated.
In order to identify the FPR-matched oracle threshold, we first narrow our search space to the average of all the thresholds tuned for accuracy.
We then search from 0.15 below this averaged threshold to 0.15 above it with a step granularity of 400.
Each threshold found by this process is applied to each seed, and from this set of thresholds we pick the threshold which best matches the FPR for each trial of tau using the deferral system.
After, we have the TPR and FPR rate for each of these thresholds FPR-matched to a tau.
The performance for each FPR-matched threshold for a specific tau is merged across seeds to get the blue x's shown in \autoref{fig:ROC} and \autoref{fig:ROCwiki}.

\section{Results on CGA-WIKI}
\label{appendix:wiki}
\begin{figure*}
    \centering
    \includegraphics[width=0.8\linewidth]{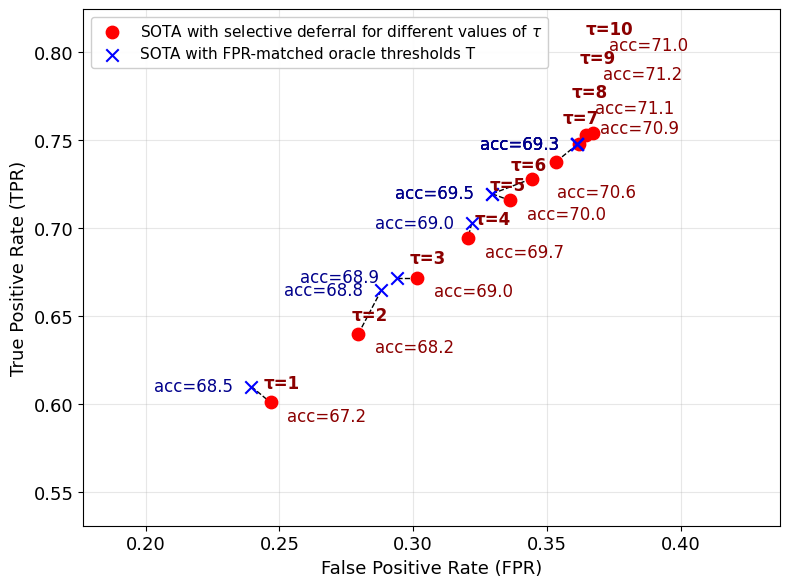}
    \caption{Decision-deferral with variable $\tau$ and matched baseline ROC performance on CGA-WIKI (averaged over five seeds).}
    \label{fig:ROCwiki}
\end{figure*}

\begin{table*}[t]
\centering
{
\begin{tabular}{lcccccc}
\toprule
Method & Acc $\uparrow$ & FPR $\downarrow$ & P $\uparrow$ & R $\uparrow$ & F1 $\uparrow$ & {H} \\
\midrule
SOTA         & \textbf{69.4} & 36.7 & 67.7 & 75.4 & \textbf{71.1} & 3.6 \\
\textbf{+ selective deferral}  & 69.2 & \textbf{35.3} & \textbf{68.1} & 71.6 & 69.6 & 3.6 \\
+ random deferral    & 68.0 & 33.6 & 67.9 & 72.7 & 70.0 & 3.5 \\
\midrule
+ simulation (average)    & 66.9 & 46.4 & 63.8 & \textbf{80.2} & 70.6 & 3.7 \\
+ simulation (majority) & 67.4 & 40.2 & 65.8 & 75.0 & 69.5 & 3.5 \\
\midrule
\midrule
with oracle threshold  & 69.5 & 33.5 & 69.0 & 72.6 & 70.3 & 3.5 \\
\bottomrule
\end{tabular}
}
\caption{Average performance on CGA-WIKI across 5 random seeds.}
\label{tab:wiki_avg_perf}
\end{table*}

Although our main focus is on the CGA-CMV dataset, we benchmark our baselines and decision-deferral method on CGA-WIKI for completeness.
As discussed in \autoref{sec:bgrelated}, since CGA-WIKI is pruned to have no personal attacks or toxicity in the body of the conversation, the recovery trajectories which decision-deferral uses to reclaim false positive errors is, by construction, sparse in CGA-WIKI.
As such, there are substantially more opportunities to initiate a deferral in CGA-CMV-Large as opposed to CGA-WIKI.

Even in these conditions, SOTA with the deferral mechanism still outperforms the SOTA on FPR, although to a lesser extent than in CGA-CMV (\autoref{tab:wiki_avg_perf}).
However, this improvement comes at a slight decrease in accuracy.
Unlike in the case of CGA-CMV, the SOTA with the deferral system is outperformed by the oracle threshold tuning, meaning that it would in principle be possible to find a threshold T which achieves similar FPR with a better precision-recall tradeoff (\autoref{fig:ROCwiki}).

\xhdr{Examples of decision-deferral on CGA-WIKI}
To concretize the FPR gains detailed in the previous section, we examine an example of a successful trigger-decision deferral in \autoref{tab:conv_wiki_win1} as done for CGA-CMV-Large in \autoref{appendix:appendixsection}.
The conversation centers the addition of an explanation of Kepler's third law, created by Speaker 1.
While Speaker 2 recognizes Speaker 1's effort, they claim that the material is not ``useful'' or ``correct.'' Speaker 1 proposes a counterargument at $t=2$, and Speaker 2 rebuts at $t=3$. Through this discussion, the conversational tension increases.
At $t=4$, Speaker 2 states that they deleted the section. 
Here, the tension $\mathcal{P} > T$.
SOTA therefore would trigger here, but because there are 10/10 calm simulations, the decision to trigger is deferred by the selective deferral system.

The following response from Speaker 1 at $t=5$ is civil, beginning with a concession---they agree with Speaker 2's stance that their added section was not useful---but they ``disagree with the rest of [their] judgement,'' defending the section.
Finally, $t=6$ concludes the conversation with a similar utterance structure as the utterance at $t=5$, beginning with an assurance that the ``example... is helpful,'' but calls the writing ``misleading.''

Although the SOTA trigger point at $t=4$ appears as an act of escalation, as the conversational context contains continuously increasing tension, the simulations predict that the next utterance will not be an escalation due to the straightforward nature of Speaker 2's comment.

\begin{table*}[h]
\centering
\small
\setlength{\tabcolsep}{5pt}
\begin{tabular}{p{0.03\linewidth} p{0.18\linewidth} p{0.51\linewidth} p{0.12\linewidth} p{0.06\linewidth}}
\hline
\textbf{t} & \textbf{Author} & \textbf{Utterance (abridged)} & \textbf{\# of calm simulations} & \textbf{$\mathcal{P}$} \\
\hline
0 & Speaker 1 & I've added a section on the relativistic case of Kepler's third law [...] & 10.0 & 0.02 \\
1 & Speaker 2 & Your hard work is admirable, but I don't think this material is useful, or even correct as currently written [...] & 8.0 & 0.29 \\
2 & Speaker 1 & In what sense are they meaningless? The Schwarzschild $t$-coordinate is observed by an observer at infinity [...] & 8.0 & 0.29 \\
3 & Speaker 2 & You've pointed out one analogy, but in other ways it fails [...] coordinates are arbitrary [...] you could pick coordinates where $P^7 \propto r^{11}$. & 9.0 & 0.44 \\
\rowcolor{lightblue}
4 & Speaker 2 & I've gone ahead and deleted the section. I hope you won't take this as confrontational [...] & 10.0 & \textbf{0.53} \\
5 & Speaker 1 & Perhaps I can accept it's not generally useful, but I disagree with the rest of your judgement [...] This fact [...] allows GPS corrections to work correctly [...] & 9.0 & 0.35 \\
6 & Speaker 2 & The GPS example you gave is helpful [...] In any case, a derivation of Kepler's third law for circular orbits in general relativity is extremely misleading to the reader [...] & 8.0 & 0.41 \\
\hline
\end{tabular}
\caption{Example thread from CGA-WIKI where decision-deferral is effective; conversation does not contain a personal attack, and thus remains calm. $\mathcal{P} := \pderail$; $\mathcal{P}$ values above $T$ are highlighted in \textbf{bold}, which are moments where SOTA will trigger. An utterance which has a trigger decision deferral is highlighted in \textcolor{mediumblue}{blue}.}
\label{tab:conv_wiki_win1}

\begin{tabular}{ c p{0.65\textwidth}}
\toprule
\textbf{$\mathcal{P}$} & \textbf{Simulated next utterance (abridged)} \\
\midrule
0.44 &
Not at all taken as confrontational. As far as your question about physicality goes, while the precise value of the constant might vary depending upon the choice of reference frame, there will always exist some values such that the ratio of revolution period squared divided by cube of separation equals said constant regardless of choice of reference frame. This allows us to say something universal about time dilation effects on planetary motion around any mass distribution. [...] \\
0.38 &
That's fine with me. Thanks again! \\
0.25 &
Not at all. My criticisms were intended to point out areas where the subject matter needed clarification rather than indicate any fault with your work. Thank you for taking my suggestions seriously. \\
0.38 &
No offense taken. Thanks! \\
0.29 &
Not at all taken personally, thanks again! I actually suspected something like that might happen after reading your initial comment, but felt it should go up regardless. Thanks! \\
0.20 &
Not confrontational at all! Thanks very much for your contribution here. [...] Could you give some idea why you think the section is incorrect? Are there particular points where you see problems? [...] \\
0.27 &
Not taken as confrontational at all! Thank you very much for being willing to admit your error rather than defending it like some people do. Also, thanks again for doing such nice work here; I wish there were more articles online about GR that took the trouble you did. [...] \\
0.41 &
Not at all! This is why we're here after all. I'll try my best to justify keeping the section in place, though I know there will probably come a time when I must give up. [...] \\
0.22 &
No offense taken at all! My main goal when writing the section was to make people interested in reading further; it certainly wasn't intended as a substitute for the real thing. [...] Thanks again for responding. 21:19 (UTC) on Sunday, August 28 2005 \\
0.32 &
Not confrontational intended nor taken. Thank you! \\
\bottomrule
\end{tabular}
\caption{
Simulated next utterances and their forecast probabilities at $t=2$, predicting comments for $t=3$.
}
\label{tab:conv_wiki_win1_sim}
\end{table*}

\section{Human Experiment Interface}
\label{appendix:humanexpscreencap}
\begin{figure*}
    \centering
    \includegraphics[width=\linewidth]{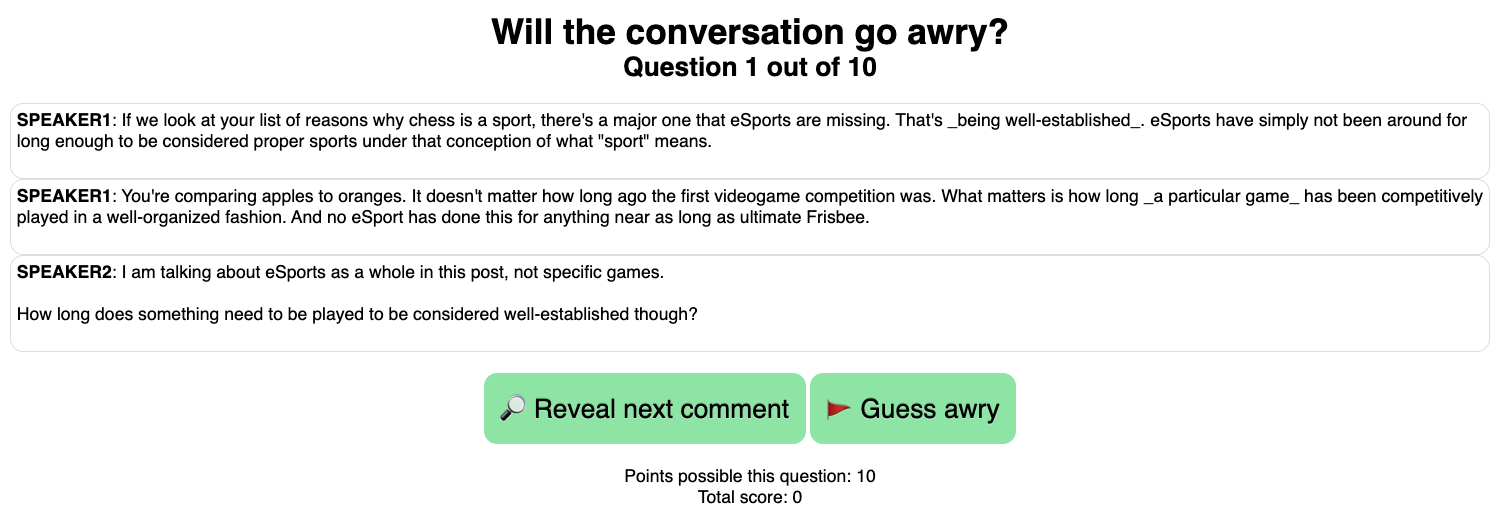}
    \caption{A screen capture of the game interface used to collect human data at a state where three utterances have been revealed so far. At each utterance, the user has a choice to reveal the next utterance or guess awry.}
    \label{fig:humaninterface}
\end{figure*}

We provide a screen capture of the interface which human participants used during our experiment in \autoref{fig:humaninterface}.

\section{Participant Characteristics}
\label{appendix:annotators}
In the novel human baseline which this work presents, we source annotations from volunteer university-level students, both undergraduate and graduate. All students are proficient in English.

\section{Implementation Details}
\label{appendix:implementation}
We describe here the practical implementation details of training the utterance simulator and forecaster used in the decision-deferral system.
We simulate utterances in these conversations using a LLaMA 3.1 8B generative model \cite{grattafioriLlama3Herd2024}, finetuned in 4-bit quantization with LoRA ($r, \alpha = 16$, no dropout or bias) on a subset of the training portion of the CGA-CMV dataset \cite{hu_lora_2022}. 
We train for 1 epochs using batch size 16, and the 8-bit AdamW optimizer \cite{loshchilov_decoupled_2019}. 
We simulate $M=10$ next utterances for each tense moment, and require that at least two-thirds of the simulations ($\tau=7$) to indicate recovery in order to defer a threshold-based triggering decision.

We adopt the SOTA model to calculate the probability of derailment, which uses a classification head on Gemma2 9B with LoRA and is trained on the train portion of the CGA-CMV-Large dataset (see \autoref{sec:bgrelated}), training for 1 epoch using effective batch size 64, learning rate $1e^{-4}$, and LoRA alpha 16.

We finetune using a cluster of 3xA6000 GPUs. Each finetuning job for simulators and forecasters on CGA-CMV-Large on a single GPU took approximately 3 GPU-hours.

\section{Summary of Contributions}
\label{appendix:summary}
\xhdr{A summary of the main paper contributions}
This paper studies conversational derailment forecasting in the Conversations Gone Awry (CGA) task, with a focus on the decision of when to trigger an alert rather than solely on estimating derailment likelihood.

The paper’s main contributions are:

\begin{itemize}
    \item A reformulation of conversational forecasting that separates belief estimation from decision making, highlighting the role of the unknown horizon in online dialogue settings.
    \item The first human baseline for the CGA task, enabled by a gamified online forecasting experiment, showing that humans achieve substantially lower false positive rates than state-of-the-art models.
    \item A simulation-based trigger deferral mechanism, inspired by human behavior, that anticipates conversational recovery and reduces false positives without sacrificing overall accuracy when layered on top of a state-of-the-art model.
\end{itemize}

Conceptually, the paper reframes conversational derailment forecasting as a sequential decision problem, offering insights that are complementary to advances in model architecture and broadly applicable to other online forecasting tasks.

\section{AI Disclosure}
\label{appendix:ai}
In this section, we disclose AI usage. AI was used for spellcheck and grammar purposes starting from a fully written text. AI was also used to automate small edits to tables in \LaTeX{}, but did not generate any text from scratch.

\end{document}